\documentclass{article}
\usepackage{spconf,amsmath,graphicx}

\usepackage{xcolor}
\usepackage{array, multirow}
\usepackage{amsfonts}
\usepackage{subcaption}

\title{Alignment Restricted Streaming Recurrent Neural Network Transducer}
\name{\begin{tabular}{c}Jay Mahadeokar, Yuan Shangguan, Duc Le, Gil Keren, Hang Su, Thong Le, Ching-Feng Yeh \\
Christian Fuegen, Michael L. Seltzer \end{tabular}}

\address{Facebook AI\\ \small \texttt{\{jaym,yuansg,duchoangle,gilkeren,suhang,tmle,cfyeh,fuegen,mikeseltzer\}@fb.com}}

\begin{document}
\maketitle

\begin{abstract}
There is a growing interest in the speech community in developing Recurrent Neural Network Transducer (RNN-T) models for automatic speech recognition (ASR) applications. RNN-T is trained with a loss function that does not enforce temporal alignment of the training transcripts and audio. As a result, RNN-T models built with uni-directional long short term memory (LSTM) encoders tend to wait for longer spans of input audio, before streaming already decoded ASR tokens. In this work, we propose a modification to the RNN-T loss function and develop Alignment Restricted RNN-T (Ar-RNN-T) models, which utilize audio-text alignment information to guide the loss computation. We compare the proposed method with existing works, such as monotonic RNN-T, on LibriSpeech and in-house datasets. We show that the Ar-RNN-T loss provides a refined control to navigate the trade-offs between the token emission delays and the Word Error Rate (WER). The Ar-RNN-T models also improve downstream applications such as the ASR End-pointing by guaranteeing token emissions within any given range of latency. Moreover, the Ar-RNN-T loss allows for bigger batch sizes and 4 times higher throughput for our LSTM model architecture, enabling faster training and convergence on GPUs.
\end{abstract}
\section{Introduction}
\label{sec:intro}
Streaming Automatic Speech Recognition (ASR) researches have made their way into our everyday products. Smart speakers can now transcribe utterances in a streaming fashion, allowing users and downstream applications to see instant output in terms of partial transcriptions~\cite{punjabi2020streaming,he2019streaming,rao2017exploring,wu2020streaming,sainath2020streaming,watanabe2017hybrid}. There is a growing interest in the community to develop end-to-end (E2E) streaming ASR models, because they can transcribe accurately and run compactly on edge devices~\cite{he2019streaming, chiu2019comparison, shangguan2019optimizing, kim2019attention}. Amongst these streaming E2E models, Recurrent Neural Network Transducer (RNN-T) is a candidate for many applications~\cite{DBLP:journals/corr/abs-1211-3711, graves2013speech}. RNN-T is trained with a loss function that does not enforce on the temporal alignment of the training transcripts and audio. As a result, RNN-T suffers from token emission delays - time from when the token is spoken to when the transcript of the token is emitted~\cite{tripathi2019monotonic,li2020developing}. Delayed emissions of tokens adversely affects user experiences and downstream applications such as the end-pointer. 

Some existing work tried to mitigate the token emission delays in streaming RNN-Ts. We introduce them in Section~\ref{sec:relatedwork}. Other works utilized semi-streaming or non-streaming models to predict better token emission time, at the cost of the overall latency of the transcripts. In this work, we propose a novel loss function for streaming RNN-T, and the resultant trained model is called Alignment Restricted RNN-T (Ar-RNN-T). It utilizes audio-text alignment information to guide the loss computation. In Section~\ref{sec:method}, we show that theoretically, Ar-RNN-T loss function is faster to compute and results in better audio-token alignment. In Section~\ref{sec:experimental_setup}, we empirically compare our proposed method with existing works such as monotonic RNN-T training~\cite{tripathi2019monotonic} on two data set: LibriSpeech and voice command. In the results section, Section~\ref{sec:results}, we show improvement in training speed and that when used in tandem with an end-pointer, Ar-RNN-T provides an unprecedentedly refined control over the latency-WER trade-offs of RNN-T models. 

\section{Related Work}\label{sec:relatedwork}
The RNN-T model consists of an encoder, a predictor and a joint network. The encoder processes incoming audio acoustic features and the predictor processes the previously emitted tokens~\cite{DBLP:journals/corr/abs-1211-3711,graves2013speech}. The joint representation of the encoder and predictor is then fed into the joint network to predict the next likely token or a `blank' symbol. Section~\ref{subsec:rnntloss} explains the loss function used to train the standard RNN-T model. Our work uses an RNN-T architecture setting similar to the RNN-T presented in~\cite{he2019streaming}. 

Several works have observed that the RNN-T loss function does not enforce alignment between audio and token emission time. Tripathi et al. noticed that the RNN-T models sometimes do not emit anything for a while before outputting multiple tokens all together~\cite{tripathi2019monotonic}. To mitigate audio-token timing misalignment, the authors proposed Mono-RNN-T models that requires the model to output at most one label at a frame. Similarly, Sak et al. proposed Neural Network Aligner (RNA)~\cite{sak2017recurrent} to train RNN models that do not output multiple labels without processing one input vector. The HAT~\cite{variani2020hybrid} variant of the RNN-T loss function is also proposed to introduce an acoustic-language modality separation in the ASR pipeline. In all these works, the exact alignment between spoken token and emitted token is implicitly modeled. Our proposed Ar-RNN-T training imposes more restriction to the audio-token and emission token alignment than the above-mentioned RNN-T variants. By leveraging audio-text alignment, we strictly enforce that the intervals between emitted tokens should be bound by a reasonable span of time.\\
\indent Zeyer et al. also explored the idea of incorporating external alignment and using approximations~\cite{sak2017recurrent} to compute the RNN-T loss function efficiently~\cite{zeyer2020new}. They reported training time and WER of non-streaming bidirectional-LSTM (BLSTM) models with their approximation framework, whereas we focus on training streaming RNN-T models, constructed with single-directional LSTMs. Our empirical results show -- similar to what~\cite{li2020developing} has discovered -- that the BLSTM-based RNN-T models with more future audio context behave differently from the single-directional LSTM-based RNN-T models in their token emission delays. We analyze the time emission delay of each token, as well as the end-pointing effects of the Ar-RNN-T training, giving users a detailed trade-off between WER and token emissions delays.\\
\indent Token emission delays are also detrimental for the RNN-T model to decide when to end-point. Li et al. trained the RNN-T to predict a special end-of-query symbol \cite{joineendpoint}, $\langle/s\rangle$, to aid the end-pointer~\cite{li2020towards}. They added penalty losses over the early and late emissions of the $\langle/s\rangle$ token to ensure that it is emitted at the right time. Their idea of early-emission and late-emission penalties is similar to Ar-RNN-T loss function. However, instead of applying these penalties only to the $\langle/s\rangle$ token, Ar-RNN-T imposes such penalty to all tokens. \\
\indent Other works introduced audio-text alignment information into the RNN-T encoder by incorporating an additional alignment restrictive loss function into model pre-training processes. In~\cite{hu2020exploring}, for instance, Hu et al. used the Cross Entropy (CE) loss function to pre-train the RNN-T encoder. Instead of imposing alignment in the pre-training step, our work directly enforces the alignment during RNN-T training. \\
\indent Besides RNN-T, there exist works sharing similar inspirations to tackle delayed emission problem for other types of model loss functions. Senior et al.~\cite{acoustic_modelling_cdctc} investigated applying constraints on the Connectionist Temporal Classification (CTC) alignment to reduce the latency in decoding CTC based models by limiting the set of search paths used in the forward-backward algorithm and Povey et al.~\cite{Povey+2016} used lattice-based technique for constraining phone emission time. Our algorithm applies to the RNN-T loss lattice to align word-piece emission times.

\begin{figure*}[h]
\begin{subfigure}{8cm}
    \centering
    \includegraphics[width=8cm]{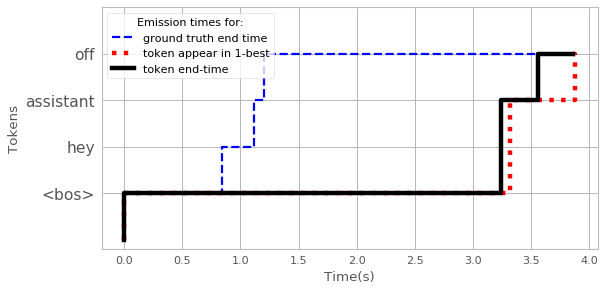}
    \caption{}\label{fig:emission}
    \vspace{-0.5em}
\end{subfigure}
\begin{subfigure}{8cm}
    \centering
    \includegraphics[width=8cm]{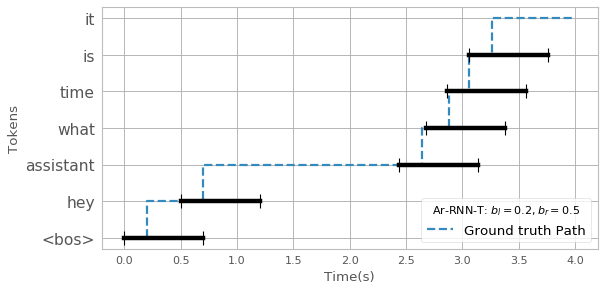}
    \caption{}\label{fig:alignment}
    \vspace{-0.5em}
\end{subfigure}
\caption{Alignment for RNN-T models. (a) shows token emission delays of a standard RNN-T on an utterance. The gap between the ``token appears in 1-best" and the ``ground truth end-time" is the token emission delay; (b) shows the Ar-RNN-T loss restricted regions for each token. \textit{$b_l$} is the left-buffer region and \textit{$b_r$} is the right-buffer measured from the spoken token end time.}
\vspace{-1.0em}
\end{figure*}

\section{Methodology} \label{sec:method}
The Recurrent Neural Network Transducer model (RNN-T)~\cite{DBLP:journals/corr/abs-1211-3711,graves2013speech} processes a sequence of acoustic input features $x = (x_1, ..., x_{\overline{T}})$ while predicting a sequence of output labels $y=(y_1, ..., y_U)$ such as characters or word-pieces~\cite{irie2019choice}. The model consist of three main components. The acoustic model sub-network processes the sequence $x$ to emit temporal representations $\overline{x} = (\overline{x}_1, ..., \overline{x}_{T})$ such that $T \leq \overline{T}$. The predictor sub-network consumes previously emitted prefix of $y$ to produce a representation summarizing the current decoding state $\overline{y} = (\overline{y}_1, ..., \overline{y}_{U})$. Finally, a joiner component combines the two representations to output a probability distribution over possible output tokens $k$, for every $\overline{x}_t$ and $\overline{y}_u$:
\begin{equation} \label{eq:joiner_out}
    z(k, t, u) = \Pr(k | \overline{x}_t, \overline{y}_u).
\end{equation}

\subsection{Standard RNN-T Loss}\label{subsec:rnntloss}
Similar to the CTC model training processes~\cite{graves2006connectionist}, RNN-T outputs a set of tokens that consist of predefined units, often word-pieces \cite{irie2019choice}, and a special blank symbol $\epsilon$, such that $z( \epsilon, t, u)$ represents the probability of not emitting any new sub-word unit given $t$, $u$.

We define an alignment $h = ((k_1, t_1), ..., (k_{T}, t_{T}))$ as a sequence of $T$ tokens emissions and define the probability of $h$ as:
\begin{equation}
    \Pr (h) = \displaystyle\prod_i z(k_i, \overline{x}_{t_i}, \overline{y}_{\text{last}(i)}),
\end{equation}
where $\text{last}(i)$ is the index of the last non-blank label among $k_1, ..., k_{i-1}$. The RNN-T model then defines the probability of a given transcription $y$ to be 
\begin{equation} \label{eq:sum_alignments}
    \Pr(y|x) = \displaystyle\sum_{h \in B^{-1}(y)} \Pr(h),
\end{equation}
where $B^{-1}(y)$ is the set of all alignments that correspond to the transcription $y$. During training, the RNN-T model attempts to maximize the probability assigned to a correct transcription $y^*$, by defining the loss as
\begin{equation}
    L = -\log(\Pr(y^*|x)).
\end{equation}
For gradient and loss computations, \cite{graves2012sequence} defines the \emph{forward variable} $\alpha(t, u)$ to be the probability of emitting $y^*_{[1:u]}$ during the first $t$ time steps:
\begin{align} \label{eq:alpha}
\begin{split}
    \alpha(t, u) =& \hspace{0.5em} \alpha(t-1, u) z(\epsilon, t-1, u) + \\
     & \hspace{0.25em} \alpha(t, u-1) z(y^*_u, t, u-1).
\end{split}
\end{align} 
Similarly, the \emph{backward variable} $\beta (t, u)$ is defined to be the probability of emitting $y^*_{[u+1:U]}$ from time step $t$ onwards.



The product $\alpha(t, u) \beta(t, u)$ is the probability of alignments that result in emitting the correct target sequence, while emitting exactly the first $u$ symbols up to time step $t$. Since $z(k, t, u)$ only affects the probability of such alignments, it holds that
\begin{align} \label{eq:grad}
\begin{split}
\frac{\partial Pr(y^*|x)}{\partial z(k, t, u)} &= \frac{\partial \alpha(t,u)\beta(t,u)}{\partial z(k, t, u)} \\
& =\alpha(t, u) 
\begin{cases}
\beta(t, u+1) \quad \text{if } k = y_{u+1}, \\ 
\beta(t+1, u) \quad \text{if } k = \epsilon, \\ 
0 \quad \text{otherwise}.
\end{cases}
\end{split}
\end{align}
The above equation allows the derivation of the RNN-T loss gradient with respect to the joiner outputs, which is then further propagated to compute the gradient of all network parameters.

When computing the RNN-T loss, we sum the probabilities of all possible alignments. 
In an extreme example, the loss function may consider an alignment in which all tokens are emitted at the last time step and therefore, model learns to wait before emitting tokens. Empirically, Figure~\ref{fig:emission} shows that standard RNN-T has a long delay between the token (spoken) end time and the emission time.

\subsection{Alignment Restricted RNN-T Loss} \label{sec:ARRNNT}
In order to have more control on token emissions, we propose restricting the set of alignments that are considered for optimization in Eq~\ref{eq:sum_alignments}. 
More formally, we augment the transcription tokens $y_1, ..., y_U$ with additional alignment labels $a_1, ..., a_U$. Alignment labels can be obtained from a bootstrap hybrid ASR model, as described in Section~\ref{sec:labels}. In addition, we define two hyper-parameters $b_l$ and $b_r$, which determine the left and right boundaries of valid alignments (see Figure~\ref{fig:alignment}). An alignment $h = (k_1, t_1), ..., (k_T, t_T)$ is considered valid by the Ar-RNN-T loss if it results in the correct transcription $y^*$, and for every $i$ such that $k_i \neq \epsilon$, meaning $k_i = y_u$ for some $u$, it holds that $a_u - b_l \leq t_i \leq a_u + b_r$. We define the valid time ranges for $u$ to be $(v_{lu}, v_{ru})$ such that $v_{lu} = a_u - b_l$ and $v_{ru} = a_u + b_r$.

The above formulation also allows a convenient computation of the loss value and the gradient. We define 
\begin{align}
\begin{split}
    \overline{z}(y_{u+1}, t, u) &= 
    \begin{cases}
    z(y_{u+1}, t, u) \text{ if } v_{lu} \leq t \leq v_{ru}, \\
    0 \text{ otherwise}.
    \end{cases} \\
    \overline{z}(\epsilon, t, u) &= z(\epsilon, t, u).
\end{split}
\end{align} 
We replace $z$ in Equation \ref{eq:alpha} for the forward $\alpha(t, u)$ computation with $\overline{z}$, and likewise for the backward $\beta(t,u)$ computation. Once the forward and backward variables are replaced with their alignment restricted counterparts, the computation of the loss and gradient in Eq. \ref{eq:grad} is identical to the standard RNN-T loss case. 

\subsection{Alignment Label Generation} \label{sec:labels}
To obtain alignment labels $a_1, ..., a_U$ we perform forced alignment using a traditional hybrid acoustic model trained using cross-entropy criterion on chenone targets~\cite{le2019senones}. Chenone model's alignment contains the silence token (SIL), which we ignore during ArRNN-T training because it is equivalent to moving along the time-dimension of the RNN-T lattice without emitting new tokens. The hybrid model provides word level start and end times $(s_j, e_j)$ for each word $w_j$ in the transcription. In contrast, our RNN-T models use word-pieces.
 We propose two strategies for obtaining word-piece alignments. The first strategy ($AS_1$) is to simply use $a_i = e_j$ for all $i, j$ such that the word-piece $y_i$ is part of the word $w_j$. The seconds strategy ($AS_2$) is to evenly split the emitted $(s_j, e_j)$ between the word-pieces that belong to the appropriate word, in the following manner: assume the word $w_j$ is comprised of $r'$ word-pieces $y_i, ..., y_{i+r'-1}$. We then set for every $1 \leq r \leq r'$: $a_{i+r-1} = s_j + \frac{r}{r'}(e_j - s_j)$. 
 Section \ref{sec:experimental_setup} contains results with both strategies. 

\subsection{Training Optimizations} \label{sec:train_optimization}
\subsubsection{Optimizing Memory}
Let $B$ denote the batch size containing training samples $S = \{s_1, s_2 ... s_B\}$; $T$, $U$ be the maximum number of encoder time-steps and output symbols for any sample $s_{i}$; $D$ be the number of word-piece targets. Our baseline implementation uses \emph{function merging} memory optimizations described in \cite{li2019improving} which fuses the loss with softmax. Implementation wise for R-NNT loss we must compute and store $\alpha(t, u)$, $\beta(t, u)$, $z(k, t, u)$ where $k \in \{y_u, \epsilon\}$ which take $O(B\times T \times U)$ memory and $gradients(b, t, u, d)$ take $O(B\times T\times U\times D)$ memory.
 
As described in \ref{sec:ARRNNT} to compute Ar-RNN-T loss we only need to consider $\alpha(t, u)$ and $\beta(t, u)$ which fall within the restriction constraints. For an efficient implementation, during training for every sample $s_i$ we pre-compute the valid time ranges $t_{iu} \in (v_{lu}, v_{ru})$ for every $y_u \in s_i$. Let $V_{s_i}$ denote the number of valid time-steps to be considered for $s_i$. Similar to ideas described in \cite{li2019improving} Sec 3.1 we then concatenate $V_{s_i}$ for $s_i \in S$ in one dimension to obtain a $2D$ tensor of dimensions $(\sum_{i} V_{s_i} , D)$ where $\sum_{i} V_{s_i}$ is $O(B\times(T + U\times(b_l + b_r)))$. This memory optimization is applicable to the joiner and all subsequent operations including loss and it allows us to scaling the model training with 4 times the batch-size on GPU clusters. 

\subsubsection{Optimizing GPU Compute}
A naive implementation that uses valid tensor locations $t_{iu}$ can be achieved using slice operation in a deep-learning library like Pytorch \cite{NEURIPS2019_9015}. To further improve throughput, we implement a custom join operation that invokes $O(\sum_{i} V_{s_i} \times D)$ cuda threads to update the joined output tensor for forward operation, each thread requires $O(B+U)$ time to execute using pre-computed indexes. For backward operation we invoke $O(B \times (T + U) \times D)$ cuda threads to update encoder and predictor gradients. We perform similar optimizations while computing $z(k, t, u)$ values inside the loss. Overall combined with an increased batch size, this optimization enables us to \emph{scale training throughput by 4 times} using the 37M parameter LSTM architecture described in \ref{subsec:librispeech}. 

Other optimization techniques like mixed precision training \cite{ott2019fairseq} are known to improve training throughput, but come at a cost of sacrificing accuracy or training stability. The proposed approach does not involve any approximations, and can be combined with mixed-precision training for added throughput improvements.

\section{Experimental Setup}
\label{sec:experimental_setup}

We trained and evaluated Ar-RNN-T models on a publicly available data set, the Librispeech~\cite{panayotov2015librispeech} corpus, as well as on an in-house, voice-search based corpus. 
\subsection{Librispeech}
\label{subsec:librispeech}
The Librispeech corpus contains 970 hours of labeled speech. We extract 40-channel filterbanks features computed from a 25ms window with a stride of 10ms. We apply
spectrum augmentation~\cite{park2019specaugment} with mask parameter (F = 27), and ten time masks with maximum time-mask ratio (pS = 0.05), and speed perturbation (perturbation ratio {0.9, 1.0, 1.1}). 

We implemented an RNN-T model architecture similar to that in~\cite{he2019streaming}. The encoder consists of 8 layers of LSTMs with Layer Norm with 640 hidden units, with a two times time dilation step after the second and fourth  LSTM layers. The decoder consists of 2 layers of LSTM with 256 hidden units. Both encoder and decoder project embeddings of 1024 dimensions. The joint layer consists of a simple DNN layer and a softmax layer, predicting a word-piece output of size 4096. 

We experiment with the amount of model look-ahead for the LSTM cells in the RNN-T. Model look-ahead is the amount of future context (or audio frames) that are provided as input to the model. Standard LSTM layers do not require any model look-ahead. We stack 10 frames with a stride of 1 as input to the standard LSTM based RNN-T encoder to providing a total of 100ms audios per decoding step. When latency-controlled Bidirectional LSTM (LC-BLSTM) cells~\cite{zhang2016highway} are used to build the RNN speech encoder~\cite{le2019senones}, it operates on 1.28s chunks with 200ms look-ahead. Bidirectional LSTM (BLSTM) cells have access to the entire incoming audio (thus maximum look-ahead). Thus, BLSTM based RNN-T models are non-streaming.

\subsection{Voice Command}\label{subsec:smartspeaker}
The voice command data set combines two sources. The first source is in-house, human transcribed data recorded via mobile devices by 20k crowd-sourced workers. The data is in the voice assistant domain, and is anonymized with personally identifiable information (PII) removed. We distort the collected audio using simulated reverberation and add randomly sampled additive background noise extracted from publicly available videos. The second source came from 1.2 million voice commands (1K hours), sampled from production traffic of smart speakers. The content of the voice commands include communication, media or music requests etc, with PII removed, audio anonymized and morphed. Speed perturbations~\cite{ko2015audio} are applied to this dataset to create two additional training data at 0.9 and 1.1 times the original speed. We applied distortion and additive noise to the speed perturbed data, resulting in a corpus of 38.6M utterances (31K hours) in total. 

Our evaluation data consists of 5K hand-transcribed anonymized utterances from volunteer participants in Smart speaker's in-house pilot program. To measure end-pointing performance thoroughly, we append 2 second silence to the end of all utterances in the voice command evaluation dataset to mitigate no end-point circumstances. The end-pointing locations are manually annotated by transcribers.  

The RNN-T model for voice command has the same architecture described in Section~\ref{subsec:librispeech}. The targets for the RNN-T training are word-pieces from a pre-trained sentence piece model~\cite{kudo2018sentencepiece}. 

\subsection{Token End Time and Finalization delay}\label{subsect:tokendelaydefine}
We analyze the delay in the emission time of ASR words during inference from the actual time when the word has been spoken in audio. 
We described the token timing alignment in Section~\ref{sec:labels}.
We define \emph{Ground Truth End Time} ($ET_{gt}$) as the end time of token as per the alignment using a hybrid ASR model. 

During inference, a breadth-first beam search algorithm explores several paths in the RNN-T decoding lattice, where each path is a sequence of ASR tokens (including blanks). Each such path can define an alignment for the predicted hypothesis text and audio.
We define \emph{ASR Token end time} ($ET_{asr}$) to be the time when the token / word was emitted according to the most probable path encountered during RNN-T decoding. Note that we use $ET_{asr}$ for downstream applications like static end-pointing.
%
We also define the \emph{Token Finalization time} ($FT_{asr}$) as the first time when the token appeared in 1-best partial hypothesis.

Given above definitions, we define the token end-time delay ($ED$) as  $ED = ET_{asr} - ET_{gt}$ and token finalization delay ($FD$) as $FD = FT_{asr} - ET_{gt}$.

\subsection{ASR End-pointing}
To compare to previous work~\cite{li2020towards}, we also study the impact of Ar-RNN-T training on the behavior of the end-pointer, which determines if the user has stopped speaking during inference. Accurate and timely end-pointing improves user experience while early or aggressive end-pointing may lead to the user being cut-off and deletion errors in the ASR hypotheses.

\subsubsection{End-pointers}
We study the behaviors of streaming RNN-T combined with the following end-pointing solutions:

\textbf{Static End-pointer} is a module that makes end-point decisions by looking at 1-best ASR partial hypothesis. An end-point decision is made when the length of trailing silence ($ET_{asr}$) exceeds the static threshold $T_{static}$.

\textbf{Neural End-pointer} (NEP) is a neural-network which takes in audio features as inputs and predicts the end-of-query (eoq) directly, as described in~\cite{shannon2017improved}. The neural end-pointer uses parameters $alpha_{eoq}$ and $T_{eoq}$ to determine an end-point, i.e. end-pointing 
when $P(eoq) >= alpha_{eoq}$ for at-least $T_{eoq}$ consecutive milli-seconds.

As described in~\cite{li2020towards},~\cite{joineendpoint} we can also train a RNN-T system to predict end of sentence ($EOS$) symbol.

\textbf{E2E End-pointer} uses the probability of $EOS$ symbol to make end-pointing decisions. More concretely, we use tunable parameters $alpha_{e2e}$ and $T_{e2e}$ and make end-point decisions only when the $P(eos) >= alpha_{e2e}$ for at-least $T_{e2e}$ consecutive milli-seconds.

\subsubsection{End-pointing metrics}

For applications that require real-time feedback, an ideal end-pointer should end-point as soon as a user stops speaking, so that results can be immediately processed to interact with the user. Since we evaluate our solution offline, we track \textbf{Endpoint Latency}~\cite{shannon2017improved} which is defined as the difference between the time end-pointer makes end-pointing decision and the time user stops speaking as annotated by transcriber. To study the distribution of end-pointing decisions, we track Average latency ($L_{avg}$) and latency at 90 percentile ($L_{p90}$).

To measure the quality of end-pointing solutions, we track two more metrics. Early cutoff percentage (Early Cut\%) is the percentage of utterances in our  which end-pointed earlier than the hand-labelled end time. No end-point percentage $NoEP_{pct}$ is the percentage of utterances where end-pointer failed to end-point at all. For optimal user experience, we aim to minimize Early Cut\% and $NoEP_{pct}$.

\section{Results}
\label{sec:results}

\subsection{LibriSpeech}
\label{ssec:finalization_delay}

\begin{table}[h]
\centering
\begin{tabular}{|p{0.12cm}| p{1.0cm} |p{1.2cm}|p{0.8cm} p{0.8cm}| p{0.9cm} p{0.9cm} |}\hline
    &\textbf{RNN-T Loss} & \textbf{\scalebox{0.8}[0.8]{Encoder}} & \textbf{Avg $ED$} & \textbf{Avg $FD$} & \textbf{Test-clean WER} & \textbf{Test-other WER} \\\hline
    A & &\scalebox{.8}[.8]{BLSTM} & -0.26 & N/A & 2.74 & 6.9 \\ 
     B & \scalebox{.8}[.8]{Standard} &\scalebox{.8}[0.8]{LC-BLSTM} & -0.13 & N/A & 3.3 & 8.4 \\
     C & &\scalebox{0.8}[0.8]{LSTM} & 0.26 & 0.33 & 4.66 & 11.7 \\ \hline
     
     D &\scalebox{.8}[.8]{Mono} &\scalebox{0.8}[0.8]{LSTM} & 0.23 & 0.27 & 4.69 & 12.0 \\ \hline
\end{tabular}
\caption{WERs and token emission delay trade-offs for different encoders trained with standard RNN-T on LibriSpeech. See Section~\ref{subsect:tokendelaydefine} for $ED$ and $FD$ definitions.}
\label{table:librispeech_result_encoders}
\vspace{-1.0em}
\end{table}

\begin{table}[h]
\centering
\begin{tabular}{| p{0.2cm} |p{1.0cm} | p{0.5 cm} |p{1.0cm} p{1.0cm}| p{0.8cm} p{0.8cm} |}\hline
       & \textbf{Align-ment} & \textbf{ $b_r$ }  & \textbf{Avg $ED$} & \textbf{Avg $FD$} & \textbf{Test-clean WER} & \textbf{Test-other WER} \\\hline
     D & $AS_1$ & 15  & 0.29 & 0.36 & 4.8 & 11.85 \\\hline
     E &        & 15  & 0.26 & 0.33 & 4.61 & 11.89 \\ 
     F & $AS_2$ & 10  & 0.17 & 0.25 & 4.91 & 12.7 \\ 
     G & & 5  & 0.07 & 0.16 & 6.38 & 15.01 \\ 
     H & & 0  & -0.01 & 0.13 & 9.31 & 19.53 \\ \hline
\end{tabular}
\caption{WERs and token emission delays on LibriSpeech for LSTM-based RNN-T models trained with Ar-RNN-T loss. $D$ is trained using alignment strategy $AS_1$ (assign word end time to all sub-words) while $E$-$H$ are trained with $AS_2$ (equally split word time among sub-words)}
\label{table:librispeech_result}
 \vspace{-0.5em}
\end{table}

\begin{table*}[h!] 
\centering
\begin{tabular}{|p{1.8cm} |p{0.8cm} | p{1.1cm} | p{0.8cm} p{0.8cm} |p{0.8cm} p{0.8cm} |p{0.8cm} p{0.8cm}|p{0.8cm} p{0.8cm} |}\hline
     \textbf{Model} & \textbf{EOS} & \textbf{End-pointer}& \textbf{WER} & \textbf{DEL} & \textbf{Early Cut\%} & \textbf{No EP\%}
     & \textbf{Avg ED} & \textbf{Avg FD} & \textbf{$L_{avg}$} & \textbf{$L_{P90}$}\\
     \hline

    {\texttt{RNN-T}} & N
     & Static & 7.5 & \textbf{4.1} & 1.42 & 0.33 & 0.60 & 0.74 &1604& 1840   \\ 
     \cline{3-7}\cline{10-11}  & &NEP & 18.5 & 15.5 & 2.06 & 0.33 & & & 692 & 1120   \\\hline
     
     \texttt{Mono-RNN-T}& N & Static & 7.6 & 4.5 & \textbf{1.25} & 0.43 & 0.46 &  0.52 & 1462 & 1695  \\ 
     \cline{3-7}\cline{10-11} & &NEP  & 13.2 & 10.3 & 1.77 & 0.68 & &  &609 & 1040 \\
     \hline
      \texttt{Ar-RNN-T}&N &Static & 7.8 & 4.2 & 2.53 & 0.14 & \textbf{0.19} & 0.31 & 1228 & 1400 \\ \cline{3-7}\cline{10-11}
      & &NEP  & 8.3 & 4.9 & 3.19 & 0.37 && &546 & 990 \\ 
     \hline
     \hline
    \texttt{Ar-RNN-T}&Y & E2E  & \textbf{7.4} & \textbf{4.1} & 2.31 & \textbf{0.12}  & \textbf{0.19} & \textbf{0.26} & 393 & 530 \\ \cline{3-7}\cline{10-11}
     \hline
\end{tabular}
\\
\caption{Comparing RNN-T with Mono-RNN-T and Ar-RNN-T with different end-pointing models on Voice Command dataset. We show WER, deletion rate, early cutoffs, percentages with no end-pointing, latency of end-pointing alongside the Avg $ED$ and Avg $FD$.}
\label{table:portal_endpoint_result}
\vspace{-1.5em}
\end{table*}

\begin{figure}[h!]
    \includegraphics[width=0.48\textwidth]{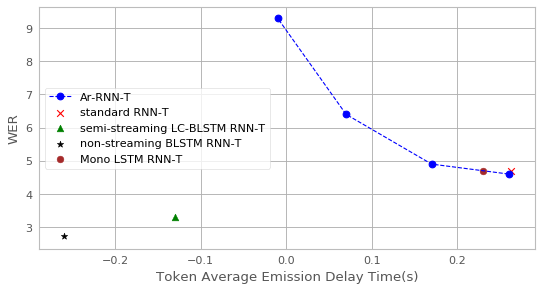}
    \caption{On LibriSpeech, Ar-RNN-T models allow refined control over the trade-offs between WER and average token end-time delays (ED).}
    \label{fig:wer-token}
     \vspace{-1.5em}
\end{figure}

As described in Table~\ref{table:librispeech_result_encoders} and Figure~\ref{fig:wer-token}, as the WER is better the more model look-ahead is provided and hence BLSTM outperforms LC-BLSTM, which in turn is better than LSTMs. More interestingly, we also study token end-time delay ($ED$) for these models and notice that LSTM model has much higher $ED$ compared to BLSTM models. The BLSTM model's Avg $ED$ is negative, which means the most probable path emission times on average are earlier than ground truth end times $ET_{gt}$. This is likely because RNN-T loss is able to use the future context built into model and confidently emit tokens earlier. However, for LSTM model the Avg $ED$ and Avg $FD$ is 263ms and 333ms respectively indicating that the model learns to wait until it sees future frames before it confidently emits the ASR tokens.
 
 For Ar-RNN-T, we run experiments using a fixed $b_l$ of 0 and sweep $b_r$. We observe that Ar-RNN-T with $b_r=15$ has better WER than the standard RNN-T. WER degrades as we impose more constraints on the buffer size. As expected we are also able to limit the emission delays with $b_r$. We plot the WER vis-à-vis average $ED$ trade-off in Figure~\ref{fig:wer-token}. Ar-RNN-T, with a configurable left-right buffer, provides a knob for researchers to tune and slide their model along the trade-off curve.

\subsection{Voice Command and End-pointing}
\label{ssec:endpointing}
On the voice command dataset, we first train an RNN-T model as described in \ref{subsec:smartspeaker}. We then fine-tune the RNN-T model with Ar-RNN-T and Mono-RNN-T losses; we also fine-tune the RNN-T model with Ar-RNN-T loss and $EOS$ symbol appended to targets for comparison. Ar-RNN-T models are trained with $b_l$=0 and $b_r$=10. For static end-pointing we use $T_{static} = 1sec$. Both the neural end-pointer (NEP) and the E2E end-pointers use static end-pointer as a fall-back solution especially when background noise is excessive. 

As shown in Table \ref{table:portal_endpoint_result}, the baseline RNN-T model with the static end-pointer has high $L_{avg}$(1604ms) and $L_{P90}$(1840ms), mainly due to delayed token emissions. Moreover when we use a neural end-pointer we see a significant increase in WER (7.5\% to 18.53\%) although the early cutoffs do not increase. This is mainly because stand-alone neural end-pointer cuts off utterances before the RNN-T model emits trailing tokens due to emission delays which leads to an increase in the deletion error (4.1\% to 15.5\%). Ar-RNN-T solves this problem as shown in row 6 of Table~\ref{table:portal_endpoint_result}, and we get WER 8.28 and $L_{avg}$ 546ms. However $L_{P90}$ is still high (990ms) mainly because static end-pointer kicks in for these 90+ percentile utterances. 

Similar to~\cite{joineendpoint}, we also train the RNN-T model with an $EOS$ symbol, but we observed that the $EOS$ symbol is extremely delayed and the model tends to emit it towards the end of audio. \cite{li2020towards} solves this problem by imposing early / late penalty on $EOS$. With Ar-RNN-T, we require no such penalty for timely $EOS$ emission. With E2E end-pointer Ar-RNN-T can achieve best trade-offs for $L_{avg}$ (393ms) and WER (7.44). Moreover, the E2E end-pointer with Ar-RNN-T model significant improves the $L_{P90}$ to 530ms. 

\section{Conclusion}
\label{sec:conclusion}
In this work, we present a detailed analysis of RNN-T model's token emission delays and its impact on the downstream applications. We propose a modification to the RNN-T loss that uses alignment information to restrict the paths being optimized during training. We call our solution Alignment Restricted RNN-T and show that we can control the token delays for RNN-T models systematically using tunable parameters during training, while also significantly improving training throughput. Using our proposed solution, we show that we can improve the accuracy of downstream applications such as the ASR end-pointing system and significantly reduce latency and early cut-offs.


\bibliographystyle{IEEEbib}
\bibliography{refs}

\end{document}